%% file: neurips_2025.tex
\documentclass{article}

\usepackage[preprint]{neurips_2025}
\usepackage{amsmath}
\usepackage[utf8]{inputenc} 
\usepackage{graphicx} 
\usepackage{dsfont}

\usepackage[dvipsnames,table,xcdraw]{xcolor} 
\usepackage{amssymb}  
\usepackage{amsthm}  
\usepackage{booktabs} 
\usepackage{tikz}
\usepackage[colorlinks=true,allcolors=teal]{hyperref}

\usepackage[capitalize,noabbrev,nameinlink]{cleveref}
\usepackage{mathtools}

\newcommand{\cP}{\mathcal{P}}

\theoremstyle{plain}
\newtheorem{axiom}{Axiom}

\newtheorem{theorem}{Theorem}[section]
\newtheorem*{theorem*}{Theorem}
\newtheorem{lemma}[theorem]{Lemma}

\newtheorem{othertheorem}{othertheorem}[section]

\newtheorem{proposition}[othertheorem]{Proposition}
\newtheorem{assumption}[othertheorem]{Assumption}

\newtheorem{definition}[theorem]{Definition}
\newtheorem{example}{Example}

\usepackage{bbold}

\usepackage[utf8]{inputenc} 
\usepackage[T1]{fontenc}    
\usepackage{hyperref}       
\usepackage{url}            
\usepackage{booktabs}       
\usepackage{amsfonts}       
\usepackage{nicefrac}       
\usepackage{microtype}      
\usepackage{xcolor}         
  \hypersetup{
  colorlinks,
  citecolor=blue!50!black,
  linkcolor=blue,
  urlcolor=black}

\title{Theoretical Tensions in RLHF: Reconciling Empirical Success with Inconsistencies in Social Choice Theory}

\author{%
  Jiancong Xiao$^1$, Zhekun Shi$^2$, Kaizhao Liu$^2$, Qi Long$^1$\thanks{Corresponding Authors}\ , and Weijie J. Su$^{1*}$\\
  $^1$University of Pennsylvania, $^2$Peking University\\
  \texttt{jcxiao@upenn.edu, \{zhekunshi,mrzt\}@stu.pku.edu.cn}\\ \texttt{qlong@upenn.edu, suw@wharton.upenn.edu} \\
}

\begin{document}

\maketitle

\begin{abstract}
Despite its empirical success, Reinforcement Learning from Human Feedback (RLHF) has been shown to violate almost all the fundamental axioms in social choice theory---such as majority consistency, pairwise majority consistency, and Condorcet consistency. This raises a foundational question: why does RLHF perform so well in practice if it fails these seemingly essential properties? In this paper, we resolve this paradox by showing that under mild and empirically plausible assumptions on the preference profile, RLHF does satisfy pairwise majority and Condorcet consistency. These assumptions are frequently satisfied in real-world alignment tasks, offering a theoretical explanation for RLHF’s strong practical performance. Furthermore, we show that a slight modification to the reward modeling objective can ensure pairwise majority or Condorcet consistency even under general preference profiles, thereby improving the alignment process. Finally, we go beyond classical axioms in economic and social choice theory and introduce new alignment criteria---\emph{preference matching}, \emph{preference equivalence}, and \emph{group preference matching}---that better reflect the goal of learning distributions over responses. We show that while RLHF satisfies the first two properties, it fails to satisfy the third. We conclude by discussing how future alignment methods may be designed to satisfy all three.
\end{abstract}

\section{Introduction}

Large Language Models (LLMs), including recent advancements such as ChatGPT-4o~\citep{openai2023gpt4} and Claude-3.7 Sonnet~\citep{anthropic2024claude}, have exhibited strong performance across a broad spectrum of tasks—ranging from code synthesis and data interpretation to mathematical reasoning and logical inference~\citep{bubeck2023sparks,chowdhery2023palm,touvron2023llama,team2023gemini}. These capabilities have accelerated their deployment in domains that were traditionally considered to require human judgment and had been viewed as resistant to automation~\citep{eloundou2024gpts}.

A key factor driving this widespread adoption is the development of methods for aligning LLM behavior with human expectations and values. Among these, Reinforcement Learning from Human Feedback (RLHF)~\citep{Ouyang2022,casper2023open,dong2024rlhf} has emerged as the dominant paradigm. RLHF typically involves training a reward model based on human preferences, often modeled using the Bradley--Terry (BT) framework~\citep{bradley1952rankanalysis}. The reward model assigns scalar scores to candidate responses, with higher scores indicating greater human preference. These scores guide the subsequent fine-tuning of the LLM, encouraging it to generate outputs that align more closely with human-annotated judgments.

However, the BT model imposes a strong assumption on the structure of human preferences. In particular, it cannot capture preference profiles that exhibit Condorcet cycles—a phenomenon where preferences are cyclic and non-transitive, such as when the majority prefers $y_1 \succ y_2$, $y_2 \succ y_3$, and $y_3 \succ y_1$. This is known as the Condorcet paradox.\footnote{Even transitive preference profiles may fail to be represented by a BT model.} In such cases, the reward modeling objective only approximates the true preference structure by fitting the best possible BT representation. This naturally leads to the question: how does the reward behave under non-BT preferences?

Answering this question requires revisiting classical results in social choice theory \citep{brandt2016handbook}, which studies how to aggregate individual preferences into a collective decision. A growing body of research has revealed a striking tension: despite the empirical success of RLHF, its reward modeling step fails to satisfy nearly all foundational axioms in social choice theory, including majority consistency, pairwise majority consistency, and Condorcet consistency~\citep{noothigattu2020axioms,siththaranjan2024distributional,ge2024axioms,maura2025jackpot,liu2025statistical}. 

To illustrate, we focus on Condorcet consistency and defer discussion of the other axioms to Section~\ref{sec:consistency}. A function is said to be Condorcet consistent if it always selects the Condorcet winner---defined as a response that is preferred by a majority of labelers over every other response---whenever such a winner exists. However, when the preference profile does not conform to the BT model, even if a Condorcet winner exists, the reward model may fail to assign it the highest score. This raises a central puzzle:
\begin{center}
    \emph{Why does RLHF perform well in practice, even though it violates these essential axioms?}
\end{center}
In this work, we begin by addressing this question. We show that under mild and empirically realistic assumptions on the preference data, RLHF satisfies both pairwise majority consistency and Condorcet consistency. In practice, each comparison is usually labeled by at most one labeler, and cyclic preferences are permitted. This gives rise to a specific structure in the preference profile: it is generally non-BT, may contain cycles, and each pairwise comparison yields a binary preference (0 or 1). When learning from such data, the reward model can correctly identify the Condorcet winner—if one exists—and assign it the highest score. This observation provides a theoretical explanation for the strong empirical performance of RLHF.

Secondly, inspired by this insight, we observe that binary preferences play a crucial role in effective LLM alignment. We consider a minimal modification to the reward modeling objective in scenarios where each comparison is labeled by more than one annotator. Instead of treating each comparisons separately, one can aggregate them and assign a binary preference based on majority vote. With this modification, the resulting reward model satisfies majority consistency, pairwise majority consistency, and Condorcet consistency. The key technical ingredient is that this formulation causes the reward modeling step to implicitly implement the Copeland rule---a classical voting mechanism in social choice theory known to satisfy these desirable axioms (see e.g., \citet{gaertner2009primer}). Notably, in many public datasets and in the labeling procedures used in the development of modern LLMs, this approach coincides with standard RLHF. Thus, our result not only provides a theoretical justification for current practices but also offers a principled guideline for future alignment strategies.

Finally, we move beyond classical frameworks in economics and social choice theory to propose new axioms tailored to LLM alignment. While economists have focused on full rankings~\citep{arrow2012social} and political scientists on selecting a winner~\citep{kelly1974voting}, generative models prioritize producing a distribution over responses. Existing axioms remain relevant but are insufficient for capturing the goals of LLM alignment. To address this gap, we introduce three distributional axioms: \emph{preference matching, preference equivalence, and group preference matching}. The first two aim to preserve preference ratios if possible, while the third ensures fairness across subgroups in non-BT, potentially cyclic, preferences. Our main theoretical results are that the target distribution is well-defined, exists, and is unique. Our results show that while RLHF satisfies the first two, it fails the third, and we offer guidance on how future methods may address this.
\section{Preliminaries: RLHF and Condorcet Paradox}
\label{Sec:pre}
\paragraph{Preliminaries of RLHF.} RLHF is an approach to align LLMs with human preferences \citep{ouyang2022training}. The standard LLM training pipeline typically involves three stages: (1) supervised fine-tuning (SFT), (2) reward modeling, and (3) policy optimization via RLHF fine-tuning. This paper primarily focuses on the reward modeling step.\footnote{Throughout the paper, we primarily refer to step (2) when discussing RLHF, unless otherwise specified.}

\paragraph{Reward Modeling.} Let $x$ be a prompt given to an LLM and $y$ be the output response. After the SFT step, the SFT model is prompted with prompts $x$ to produce pairs of answers $(y_1, y_2) \sim \pi_{\textnormal{SFT}}$. After that, human labelers express their preferences for one of the two answers, labeled as $y_\textnormal{w} \succ y_\textnormal{l}$, where $y_\textnormal{w}$ and $y_\textnormal{l}$ denote the winner and loser among $(y_1, y_2)$, respectively. The preference model is assumed to be the BT model \citep{bradley1952rankanalysis}, which can be written as
\begin{equation}
\label{eq:BT}
\mathcal{P}(y_1 \succ y_2|y_1, y_2,x)=\frac{\exp(r(x,y_1))}{\exp(r(x,y_1))+\exp(r(x,y_2))}.
\end{equation}
\noindent By collecting an offline dataset of comparisons $D_\textnormal{comp}$, we
can learn a reward via maximum likelihood:\begin{equation}
\label{eq:rewardloss}
    \min_r - \mathbb{E}_{(x,y_\textnormal{w}, y_\textnormal{l})\sim D_\textnormal{comp}}[\log(\sigma(r(x,y_\textnormal{w})-r(x,y_\textnormal{l})))],
\end{equation}
where $\sigma(u) := 1/(1+ \exp(-u))$ is the logistic function. For notational simplicity, we omit the dependence
on the prompt $x$ when it does not cause confusion in the following discussion.

Let \(\mathcal{Y}=\{y_1,\ldots,y_n\}\) be a set of \(n\) distinct responses, which is also refereed to as \emph{candidates} in social choice theory, and let \(\mathcal{U} = \{1, \ldots, m\}\) be a set of \(m\) labelers, known as \emph{voters}.
\begin{definition}[Preference Profile]
\label{def:preference_profile}
Given any two distinct responses $y, y' \in \mathcal{Y}$, a human labeler $i \in \mathcal{U}$ may express a preference for either $y \succ y'$ or $y'\succ y$. The preference of labeler $i$ is represented as a set of binary comparisons over the responses, denoted by $\tau_i$\footnote{For simplicity, we assume strict preferences without ties. Ties, if present, can be resolved arbitrarily.}. The collective preference profile of the group is defined as $\boldsymbol{\tau} = (\tau_i)_{i \in \mathcal{U}}$.
\end{definition}

In the classical setting, referred to as the \emph{complete preference profile}, each voter provides a complete and transitive ranking; that is, $\tau_i$ corresponds to a permutation over $\mathcal{Y}$. In contrast, we also consider a \emph{generalized preference profile}, where each individual preference relation $\tau_i$ may be \emph{incomplete}—some pairs $(y, y')$ are not compared and thus not included in $\tau_i$---and \emph{non-transitive}, meaning that cycles are allowed and transitivity is not required.

We denote $\mathcal{P}(y \succ y')$ as the proportion of comparisons between $y$ and $y'$ in which $y$ is ranked above $y'$. If $\mathcal{P}(y \succ y') > 0.5$, we say that $y$ defeats $y'$ under the pairwise majority rule.

\paragraph{RLHF in General Non-BT Preference.}
The foundational assumption of RLHF is that human preferences can be modeled by a BT model, which corresponds to the binary case of the Plackett–Luce (PL) model with a single latent reward. A minimal requirement of the BT or PL framework is that the underlying human preferences are transitive. However, human preferences are not always transitive---a phenomenon known as the Condorcet cycle. Below, we provide a classical example illustrating the Condorcet paradox.
\begin{example}[Condorcet paradox \citep{gehrlein2006condorcet}]\label{ex:condorcet}
    Suppose one-third of the population prefers $y_1\succ y_2\succ y_3$, one-third prefers $y_2\succ y_3\succ y_1$, and the remaining third prefers $y_3\succ y_1\succ y_2$. In this case we have $\cP(y_1\succ y_2)=\frac{2}{3}$, $\cP(y_2\succ y_3)=\frac{2}{3}$, and $\cP(y_3\succ y_1)=\frac{2}{3}$. Therefore this preference relationship is cyclic.
\end{example}
Such a Condorcet cycle implies a contradiction in the reward values: it would require $r_1 > r_2 > r_3 > r_1$, which is impossible. Therefore, if a preference profile contains a Condorcet cycle, it cannot be represented by a BT (or PL) model. As shown in \citet{liu2025statistical}, the probability that such a cycle arises is high; for instance, with just three labelers, this probability approaches 1 polynomially as $n \to \infty$. To understand the behavior of RLHF under non-BT preferences, it is essential to analyze the optimal solution of the reward modeling objective with a general non-BT preference.

\section{Pairwise Majority Consistency and Condorcet Consistency of RLHF}
\label{sec:consistency}
Answering the above question requires delving into social choice theory, which studies how to aggregate individual preferences within a group to arrive at an optimal collective decision. We begin by briefly introducing the basic concepts of social choice theory.

\begin{definition}[Aggregation Rule]
\label{def:aggregate_rule}
An \emph{aggregation rule} (or social choice function) is a function 
\[
f: \mathcal{T}^m \to \mathcal{O}(\mathcal{Y}),
\]
where $\mathcal{T}^m$ denotes the set of all possible preference profiles over $m$ voters, and $\mathcal{O}(\mathcal{Y})$ denotes the set of strict rankings over the response set $\mathcal{Y}$.
\end{definition}
Given a preference profile $\boldsymbol{\tau} = (\tau_1, \ldots, \tau_m)$, the function $f(\boldsymbol{\tau})$ returns an aggregated ranking over $\mathcal{Y}$ that represents the collective preference of the group.
\paragraph{Reward Modeling as an Aggregation Rule.}
Based on Definitions~\ref{def:preference_profile} and~\ref{def:aggregate_rule}, the maximum likelihood estimation (MLE) of the objective of reward modeling (Equation~\eqref{eq:rewardloss}) can be interpreted as an aggregation rule: it takes the preference profile---represented as the comparison dataset $D_{\textnormal{comp}}$---as input, and outputs a reward function whose induced values define a strict total ranking over the responses.\footnote{We will separately discuss how step (3) in RLHF preserves the ranking of the reward in the appendix.}

In social choice theory, a desirable aggregation rule may satisfy one or more of the following properties.

\begin{definition}
\label{def:desirable_properties}
Let $\boldsymbol{\tau} = (\tau_i)_{i \in \mathcal{U}}$ be a preference profile over a set of responses $\mathcal{Y}$, and let $f(\boldsymbol{\tau})$ denote the aggregate ranking. An aggregation rule $f$ may satisfy the following (desirable) properties:
\begin{enumerate}
\item \textbf{Pareto Optimality:}  
If all comparisons between $y_i$ and $y_j$ that exist agree on $y_i \succ y_j$, i.e., $\mathcal{P}(y_i \succ y_j) = 1$, then the aggregate ranking must also place $y_i$ above $y_j$.

\item \textbf{Majority Consistency:}  
If there exists a response $y^*$, called the majority winner, that is ranked in the top position by more than 50\% of voters, then $y^*$ must be ranked first in the aggregate ranking.

\item \textbf{Pairwise Majority Consistency:}  
If there exists a ranking $\tau^*$, called the pairwise majority consistent ranking, such that for all $y_i, y_j \in \mathcal{Y}$, $y_i \succ_{\tau^*} y_j$ if and only if a majority of voters prefer $y_i$ over $y_j$, i.e., $\mathcal{P}(y_i \succ y_j) > 1/2$, then the aggregation rule must return this ranking, i.e., $f(\boldsymbol{\tau}) = \tau^*$.

\item \textbf{Condorcet Consistency:}  
If there exists a response $y^\star$, called the Condorcet winner, that defeats every other response in pairwise majority comparisons, i.e., $\mathcal{P}(y^\star \succ y_i) > 1/2$ for all $y_i \in \mathcal{Y} \setminus \{y^\star\}$, then $y^\star$ must be ranked first in the aggregate ranking.

\end{enumerate}
\end{definition}

Pareto optimality is a minimal requirement, ensuring that universally preferred responses are respected, and is satisfied by virtually all standard aggregation rules. A desirable aggregation rule should also satisfy at least one of the properties (2)–(4) in Definition~\ref{def:desirable_properties}. 


Unfortunately, a number of prior studies have shown that RLHF satisfies only the minimal requirement of Pareto optimality, while failing to meet any of the stronger criteria---majority consistency, pairwise majority consistency, or Condorcet consistency. \citet{noothigattu2020axioms} first showed that the MLE of the objective in reward modeling (Equation~\eqref{eq:rewardloss}) does not satisfy pairwise majority consistency. Their counterexample relies on varying the number of comparisons across different pairs of candidates. When the number of comparisons is fixed across all pairs, \citet{ge2024axioms} further proved that linear aggregation rules\footnote{A linear aggregation rule refers to the setting that the reward function is linear in its parameters.} still fail to satisfy pairwise majority consistency.

\paragraph{Borda Count Rule.} A fundamental result was established by \citet{siththaranjan2024distributional}, who showed that the MLE of the reward modeling objective implicitly implements the Borda count rule. Specifically, the Borda count of a response $y_i$ is defined as $\textnormal{BC}_i = \sum_{k \neq i} \mathcal{P}(y_i \succ y_k)$. It was shown that the MLE satisfies $r(y_i) > r(y_j) \Leftrightarrow \textnormal{BC}_i > \textnormal{BC}_j$. However, the Borda count is a classical voting mechanism that violates properties (2)–(4), satisfying only the minimal requirement of Pareto optimality.\footnote{For linear aggregation rules, even Pareto optimality may be violated \citep{ge2024axioms}.} This suggests that RLHF, when viewed through the lens of Borda-based aggregation, may struggle to faithfully align with human preferences.

\subsection{Why does RLHF Achieve Remarkable Success?}
The failure of RLHF to satisfy nearly all majority axioms suggests that it may be fundamentally limited in capturing human preferences. This result is counterintuitive, as RLHF has demonstrated strong empirical performance in aligning large language models with human feedback. This paradox raises a natural question: why does RLHF achieve such remarkable success in practice?

To answer this question, we revisit the structure of the underlying preference profile. One possible explanation is that, for each prompt, only two responses are compared---thus preventing the formation of cycles. While this assumption holds for some public pairwise comparison datasets, it does not reflect the data collection process used in training advanced LLMs such as GPT-4. As noted in \citet{ouyang2022training}: ``We present labelers with anywhere between $K = 4$ and $K = 9$ responses to rank. This produces $\binom{K}{2}$ comparisons for each prompt shown to a labeler.'' This setting involves more than three comparisons per prompt, which makes it possible for cyclic preferences to arise and for the Condorcet paradox to occur.

A key observation is that each comparison is labeled by only one labeler, as assigning two labelers per comparison would effectively double the overall annotation cost. To our knowledge, there is no requirement that a labeler must provide a strict total ranking over the set of responses for a given prompt. Based on these observations, we introduce the following assumption to reflect the practical preference labeling process.

\begin{assumption}
\label{ass:one_labeler}
For any two distinct responses $y, y' \in \mathcal{Y}$, exactly one human labeler $i \in \mathcal{U}$ provides a comparison, expressing either $y \succ_i y'$ or $y' \succ_i y$. Cyclic preferences are allowed within the comparisons provided by a single labeler. The labeler responsible for each pairwise comparison may vary across different pairs.
\end{assumption}
One might think that if each comparison is labeled by a single labeler, then no conflict arises, making this the simplest case. However, the following example demonstrates that this is not necessarily true.
\begin{example}
Consider a prompt with three responses. Consider a preference dataset only includes three pairwise comparisons: $y_1 \succ y_2$, $y_2 \succ y_3$, and $y_3 \succ y_1$. These comparisons may be labeled by the same labeler or by different labelers.
\end{example}
We can observe that Condorcet cycles can still arise. In fact, any form of cyclic preference that may appear in a general preference profile can also occur under Assumption~\ref{ass:one_labeler}. This raises the same fundamental question posed in the general setting: does RLHF satisfy the axioms in Definition~\ref{def:desirable_properties} under Assumption~\ref{ass:one_labeler}?

\begin{theorem}
\label{thm:majorityandcondorcet}
Under Assumption~\ref{ass:one_labeler}, the aggregation rule of reward modeling in Equation~\eqref{eq:rewardloss} satisfies pairwise majority consistency and Condorcet consistency.
\end{theorem}
Theorem~\ref{thm:majorityandcondorcet} highlights the ability of RLHF to preserve the overall ranking or the winner. In particular, under Assumption~\ref{ass:one_labeler}, if the preference profile contains a strict ranking or a Condorcet winner, then the MLE of the reward modeling objective will recover that ranking or winner. This result provides a theoretical explanation for the strong empirical performance of RLHF. Under this practical assumption, RLHF can effectively align with human preferences, which is not generally guaranteed in the case of arbitrary preference profiles. To conclude this subsection, we examine the Pareto optimality and majority consistency of RLHF.
\begin{proposition}
Under Assumption~\ref{ass:one_labeler}, the aggregation rule of reward modeling in Equation~\eqref{eq:rewardloss} does not satisfy Pareto optimality or majority consistency.
\end{proposition}

Regarding Pareto optimality, it is important to note that this axiom cannot, in general, be satisfied when individual preferences are allowed to be cyclic. For example, suppose a labeler provides the comparisons $y_1 \succ y_2$, $y_2 \succ y_3$, and $y_3 \succ y_1$. Although all available comparisons support $y_1 \succ y_2$, RLHF would assign equal rewards $r_1 = r_2 = r_3$, thereby violating Pareto optimality.

As for majority consistency, the notion of a majority winner becomes ill-defined under Assumption~\ref{ass:one_labeler}, since individual preferences may be cyclic, and a well-defined top-ranked response may not exist.

However, we emphasize that this limitation is not critical. In the next section, we show how RLHF can be slightly modified to satisfy the axioms---including Pareto optimality and majority consistency---under more general complete preference profiles involving multiple labelers.

\subsection{Copeland RLHF}
We now show that a slight modification to the reward modeling objective in Equation~\eqref{eq:rewardloss} enables it to satisfy the desirable properties outlined earlier. Under Assumption~\ref{ass:one_labeler}, the pairwise preference $\mathcal{P}(y_i \succ y_j)$ between any two responses $y_i$ and $y_j$ is binary---either 0 or 1---which plays a key role in the proof of Theorem~\ref{thm:majorityandcondorcet}. This observation motivates the following idea: when each comparison is labeled by more than one labeler, we should not treat each comparison separately. Instead, we aggregate the comparison and use a single data point weighted by an indicator function $\mathds{1}[\mathcal{P}(y_i \succ y_j) > 1/2]$, reflecting the majority preference. This leads to the following modified loss function:
\begin{equation}
\label{eq:rewardloss2}
    \min_r - \sum_{y_i, y_j,\ i \neq j} \mathds{1}[\mathcal{P}(y_i \succ y_j) > 1/2] \cdot \log\left(\sigma\big(r(y_i) - r(y_j)\big)\right).
\end{equation}
\begin{theorem}
\label{thm:majorityandcondorcet2}
Given a complete preference profile, the aggregation rule of reward modeling in Equation~\eqref{eq:rewardloss2} satisfies Pareto optimality, majority consistency, pairwise majority consistency and Condorcet consistency.
\end{theorem}
It is also assumed in some studies that such an aggregation label is used in the training of de facto LLMs~\citep{wang2024helpsteer,kopf2023openassistant}. If this is the case, Theorem~\ref{thm:majorityandcondorcet2} provides a theoretical justification for such RLHF practices. In this paper, however, we adopt the labeling setup described in the GPT-4 Technical Report~\citep{openai2023gpt4}.

The loss function in Equation~\eqref{eq:rewardloss2} had also been discussed in \citet{ge2024axioms}, where it is shown that the corresponding linear aggregation rule satisfies pairwise majority consistency, though it still fails to satisfy the other desirable properties. In contrast, our focus is on the MLE, which distinguishes our analysis from theirs. In their discussion, they further highlighted that the Copeland rule from social choice theory possesses several favorable properties beyond pairwise majority consistency. Motivated by this, they propose a new alignment method called lexical Copeland to emulate the Copeland rule and thereby satisfies Pareto optimality and majority consistency.

In this paper, we present a cleaner and more direct result in Theorem~\ref{thm:majorityandcondorcet2}, whose proof relies on showing that the objective in Equation~\eqref{eq:rewardloss2} implicitly implements the Copeland rule, as formally stated in Lemma~\ref{lemma:copeland}, without the need to introduce more complex methods.

\begin{lemma}
\label{lemma:copeland}
Let $w_i$ denote the total number of pairwise majority wins for response $y_i$, defined as
\[
w_i = \sum_{k \neq i} \mathds{1}[\mathcal{P}(y_i \succ y_k) > 1/2].
\]
Then, the MLE of the objective in Equation~\eqref{eq:rewardloss2} implements the Copeland rule. That is, for all $i, j$, $
r_i > r_j \;\Leftrightarrow\; w_i > w_j$.
\end{lemma}
We refer to this approach as \emph{Copeland RLHF}. Comparing standard RLHF and Copeland RLHF reduces to comparing the Borda count rule with the Copeland rule. Under the Copeland rule, if $y_i$ is a Condorcet winner—meaning it defeats all other candidates in pairwise comparisons—then it achieves the maximum number of wins, $w_i = n - 1$, and is ranked first. In contrast, under the Borda count rule, the Borda score is defined as $\textnormal{BC}_i = \frac{1}{n} \sum_{k \neq i} \mathcal{P}(y_i \succ y_k)$, and the Condorcet winner may not receive the highest score, potentially leading to a different top-ranked response.

By definition, standard RLHF and Copeland RLHF coincide under Assumption~\ref{ass:one_labeler}. However, we argue that the fact that RLHF satisfies pairwise majority and Condorcet consistency in practice is largely incidental, driven by the cost constraints that limit each comparison to a single labeler. In future settings, where developers may collect comparisons from multiple labelers, care must be taken---Copeland RLHF may provide a more principled and robust alternative.

\section{Axioms for Preserving Diverse Human Preference}
\label{sec:axioms}
In this section, we move beyond the traditional frameworks of economics and social choice theory to propose new axioms that alignment methods for LLMs should satisfy in order to preserve diverse human preferences. Economists, following the tradition of Arrow \citep{arrow2012social}, have typically focused on the full ranking of candidates, while political scientists have often emphasized the selection of a single winner \citep{kelly1974voting}. But what about LLM developers? For generative models, the distribution over responses given a prompt, denoted by $\pi(y|x)$, is more central than either the ranking of responses or the choice of a single winning response. While the axioms from economics and social choice theory remain relevant, they do not fully capture the desiderata that alignment methods for LLMs should satisfy. To this end, we introduce the concept of a probabilistic aggregation rule.\footnote{Probabilistic aggregation rules have also been studied in social choice theory, typically focusing on the probability of each candidate being selected as the winner. This differs from our setting.}

\begin{definition}[Probabilistic Aggregation Rule]
\label{def:aggregate_rule2}
A \emph{probabilistic aggregation rule} (or probabilistic social choice function) is a function 
\[
\rho: \mathcal{T}^m \to \Delta(\mathcal{Y}),
\]
where $\mathcal{T}^m$ denotes the set of all possible preference profiles over $m$ voters, and $\Delta(\mathcal{Y})$ denotes the set of distributions over the response set $\mathcal{Y}$.
\end{definition}
Given a preference profile $\boldsymbol{\tau} = (\tau_1, \ldots, \tau_m)$, the function $\rho(\boldsymbol{\tau})$ returns an aggregated distribution over the response set $\mathcal{Y}$ that captures the collective preference. Specifically, $\rho(\boldsymbol{\tau}) = (p_1, \ldots, p_n)$, where $\sum_{i=1}^n p_i = 1$ and each $p_i$ denotes the generation probability assigned to response $y_i$. 
\paragraph{Reward Modeling as a Probabilistic Aggregation Rule.}
Based on Definition~\ref{def:aggregate_rule}, the MLE of the reward modeling objective can be viewed as a probabilistic aggregation rule. Once the reward values for the $n$ responses are learned, applying a softmax—consistent with the BT or PL model—yields a distribution $\boldsymbol{p}$ over responses.\footnote{Again, we will separately discuss how step (3) in RLHF preserves the distribution induced by the reward.}

This leads to the central question: what properties should the distribution $\boldsymbol{p}=(p_1, \ldots, p_n)$ satisfy in order to faithfully preserve the human preferences encoded in $\boldsymbol{\tau}$?
\subsection{Minimal Desideratum: Preference Matching}
We begin with the simplest setting: there are only two responses, $y_1$ and $y_2$. Suppose three voters prefer $y_1$ and two voters prefer $y_2$, i.e., $\mathcal{P}(y_1 \succ y_2) = 0.6$. In this case, the majority (and Condorcet) winner is $y_1$. However, if an LLM generates only the majority winner, the preferences of the two minority voters are entirely ignored, raising concerns about fairness. A more balanced approach is to match the preferences by generating $y_1$ with probability 60\% and $y_2$ with probability 40\%. 

In more complex settings, such as when a Condorcet cycle exists, it may be impossible to match all pairwise comparisons simultaneously. Nonetheless, we propose a minimal requirement for probabilistic aggregation: if there exists a distribution that is consistent with all pairwise preferences in the profile, the aggregation rule should return such a distribution. We formalize this requirement as the following axiom.

\begin{axiom}[Preference Matching]
\label{ax:pm}
Given a preference profile $\boldsymbol{\tau}$, if there exists a distribution $\boldsymbol{p}^*=(p_1^*,\ldots,p_n^*)$, called the preference matching distribution, such that, for all $i\neq j$, $p_i^*:p_j^*=\mathcal{P}(y_i\succ y_j)$, then the probabilistic aggregation rule must return $\boldsymbol{p}^*$.
\end{axiom}
The notion of preference matching also appears in \citet{xiao2024algorithmic}, though their focus is on step (3) of RLHF, which differs from our setting. A natural question that follows is: when does such a distribution exist? It turns out that this condition is precisely characterized by BT model assumption.
\begin{theorem}
\label{thm:pm}
Given a preference profile $\boldsymbol{\tau}$, a preference matching distribution exists if and only if $\boldsymbol{\tau}$ can be represented by a BT model with a single reward, referred to as BT-embeddable.
\end{theorem}

Therefore, the preference matching axiom imposes a requirement on probabilistic aggregation only under the BT assumption. RLHF satisfies the preference matching axiom, as it recovers the BT reward when the underlying preference profile is BT-embeddable. In the following sections, we explore the desirable properties of AI alignment under non-BT preferences.
\subsection{Advanced Desideratum: Preference Equivalence}
We now step back to the classical Condorcet paradox. Consider three labelers and three responses, with the preference profile given by $\boldsymbol{\tau} = (y_1 \succ y_2 \succ y_3,\ y_2 \succ y_3 \succ y_1,\ y_3 \succ y_1 \succ y_2)$. In this case, the preferences form a Condorcet cycle. It is impossible to construct a distribution that satisfies preference matching; for example, the implied pairwise ratios would require $p_1^* : p_2^* = 2 : 1$, $p_2^* : p_3^* = 2 : 1$, and $p_3^* : p_1^* = 2 : 1$, which cannot simultaneously hold. From another perspective, the responses $y_1$, $y_2$, and $y_3$ are cyclically symmetric, and thus should be assigned equal probability, i.e., $p_1^* = p_2^* = p_3^* = 1/3$. Therefore, under non-BT preferences, it is natural to have a requirement: if two responses are equally preferred according to the aggregate preference information, they should be assigned equal generation probabilities. To define what it means for responses to be equally preferred, we begin with the notion of equivalent preference profiles.

\begin{definition}[Equivalent Preference Profile]
Let $\boldsymbol{\tau}$ and $\boldsymbol{\tau}'$ be two preference profiles over $m$ voters and $n$ responses. We say that $\boldsymbol{\tau}$ and $\boldsymbol{\tau}'$ are \emph{equivalent} if there exists a one-to-one mapping between the responses in the two profiles such that the $m$ rankings are preserved under this mapping.
\end{definition}
We are now ready to formally introduce a desirable axiom, which we refer to as \emph{preference equivalence}.
\begin{axiom}[Preference Equivalence]
Given a preference profile $\boldsymbol{\tau}$, if two responses $y_i$ and $y_j$ are equally preferred---that is, there exists a one-to-one mapping that exchanging $y_i$ and $y_j$ in all individual rankings and results in a new preference profile $\boldsymbol{\tau}'$ that is equivalent to $\boldsymbol{\tau}$---then the probabilistic aggregation rule must assign equal probabilities: $p_i^* = p_j^*$.
\end{axiom}
We use the example of the Condorcet paradox to illustrate the preference equivalence axiom. In the preference profile $\boldsymbol{\tau} = (y_1 \succ y_2 \succ y_3,\ y_2 \succ y_3 \succ y_1,\ y_3 \succ y_1 \succ y_2)$, by relabeling $y_1$ to $y_2$ and $y_2$ to $y_1$, we obtain a new profile $\boldsymbol{\tau}' = (y_2 \succ y_1 \succ y_3,\ y_1 \succ y_3 \succ y_2,\ y_3 \succ y_2 \succ y_1)$. These two profiles are equivalent under the mapping $y_1 \rightarrow y_2$, $y_2 \rightarrow y_1$, and $y_3 \rightarrow y_3$. Therefore, $y_1$ and $y_2$ are equally preferred. By applying similar mappings, we have $y_1$ and $y_3$, as well as $y_2$ and $y_3$, are also equally preferred. As a result, the probabilistic aggregation rule must assign equal probabilities: $p_1^* = p_2^* = p_3^* = 1/3$. Finally, we show that RLHF do satisfied this axiom. 
\begin{theorem}
\label{thm:rlhf_pe}
The probabilistic aggregation rule induced by reward modeling satisfies the preference equivalence axiom.
\end{theorem}

\subsection{Ultimate Desideratum: Group Preference Matching}
In the last subsection, we address a fundamental question: what should the distribution $\boldsymbol{p}^*$ be when the preference profile is not BT-embeddable? Once again, we begin with the classical Condorcet paradox. Given the preference profile $\boldsymbol{\tau} = (y_1 \succ y_2 \succ y_3,\ y_2 \succ y_3 \succ y_1,\ y_3 \succ y_1 \succ y_2)$, how can we reconcile the cyclic preferences it contains? A natural approach is to partition the voters into two subgroups without cyclic preference. The first subgroup consists of a single voter with preference profile $\boldsymbol{\tau}_1 = (y_1 \succ y_2 \succ y_3)$, while the second subgroup consists of two voters with preference profile $\boldsymbol{\tau}_2 = (y_2 \succ y_3 \succ y_1,\ y_3 \succ y_1 \succ y_2)$. To determine the probability of generating $y_1$, denoted $p_1^*$, observe that in the first subgroup, 100\% of voters prefer $y_1$, while in the second subgroup, no voter prefers $y_1$. To fairly reflect the preferences of both subgroups, it is desirable to take a weighted average of these preferences, with weights proportional to the subgroup sizes. Specifically, we compute $p_1^* = 1/3 \times 100\% + 2/3 \times 0\% = 1/3$. Applying this approach to all responses, we recover the desirable distribution: $p_1^* = p_2^* = p_3^* = 1/3$. Based on this idea, we formalize the concept of group preference matching distribution.
\begin{definition}[Group Preference Matching Distribution]
\label{def:gpmd}
Given a preference profile $\boldsymbol{\tau} = (\tau_i)_{i \in \mathcal{U}}$, suppose there exists a partition $\mathcal{U} = \mathcal{U}_1 \cup \mathcal{U}_2 \cup \cdots \cup \mathcal{U}_K$ into $K$ disjoint subgroups, where each subgroup $\mathcal{U}_k$ has a BT-embeddable preference profile $\boldsymbol{\tau}_k = (\tau_i)_{i \in \mathcal{U}_k}$ with an associated preference matching distribution $\boldsymbol{p}^*_k$. Then, the group preference matching distribution\footnote{The expression has a similar form to a traditional model known as the mixture of BT/PL models. However, the underlying concept is fundamentally different. We defer a detailed discussion to the appendix.
} is defined as
\begin{equation*}
    \boldsymbol{p}^* = \sum_{k=1}^K \frac{|\mathcal{U}_k|}{|\mathcal{U}|} \, \boldsymbol{p}^*_k.
\end{equation*}
\end{definition}
In the next two theorems, we show that the group preference matching distribution is well-defined—that is, it exists and is unique. 
\begin{theorem}[Existence of Group Preference Matching Distribution]
\label{thm:exist}
Given a complete preference profile $\boldsymbol{\tau} = (\tau_i)_{i \in \mathcal{U}}$, there exists a partition $\mathcal{U} = \mathcal{U}_1 \cup \mathcal{U}_2 \cup \cdots \cup \mathcal{U}_K$ into $K$ subgroups such that the preference profile of each subgroup is BT-embeddable. Consequently, the group preference matching distribution $\boldsymbol{p}^*$, as defined in Definition~\ref{def:gpmd}, exists.
\end{theorem}

\begin{theorem}[Uniqueness of Group Preference Matching Distribution]
\label{thm:unique}
Given a complete preference profile $\boldsymbol{\tau} = (\tau_i)_{i \in \mathcal{U}}$, suppose there exist two partitions of the voter set: $\mathcal{U} = \mathcal{U}_1 \cup \mathcal{U}_2 \cup \cdots \cup \mathcal{U}_K$ and $\mathcal{U} = \mathcal{U}'_1 \cup \mathcal{U}'_2 \cup \cdots \cup \mathcal{U}'_{K'}$, such that the preference profile of each subgroup in both partitions is BT-embeddable. Let $\boldsymbol{p}^*$ and $\boldsymbol{p}^{*'}$ be the corresponding group preference matching distributions. Then, $\boldsymbol{p}^* = \boldsymbol{p}^{*'}$.
\end{theorem}
We now proceed to define the strongest desirable property.
\begin{axiom}[Group Preference Matching]
Given a complete preference profile $\boldsymbol{\tau}$, the probabilistic aggregation rule $\rho$ must return the group preference matching distribution $\boldsymbol{p}^*$, i.e., $\rho(\boldsymbol{\tau})=\boldsymbol{p}^*$.
\end{axiom}
This represents a strong requirement for AI systems to faithfully preserve human preferences. Although we have shown that RLHF is consistent with traditional social choice theory under certain conditions, it does not satisfy this property.
\begin{proposition}
There exist complete preference profiles for which the MLE of the reward modeling objective does not recover the group preference matching distribution $\boldsymbol{p}^*$.
\end{proposition}

To conclude this section, we discuss a potential modification to the reward modeling objective that enables learning the target distribution $\boldsymbol{p}^*$. To minimize deviation from the standard formulation, we could adjust the weighting scheme in the loss as follows:
\begin{equation}
\label{eq:rewardloss3}
    \min_r - \sum_{y_i, y_j,\ i \neq j} \frac{p_i^*}{p_i^* + p_j^*} \cdot \log\left(\sigma\big(r(y_i) - r(y_j)\big)\right).
\end{equation}
Under this formulation, the MLE recovers reward values that are consistent with the target distribution $\boldsymbol{p}^*$. The remaining challenge is how to estimate $\boldsymbol{p}^*$ in practice. Fortunately, it is not necessary to explicitly identify preference subgroups. A simple and effective approximation is to set $p_i^* \approx \#(y_i\ \text{ranked first}) / m$, where $m$ is the total number of voters. We leave a more detailed analysis of this approximation to the appendix.

\section{Related Work}
A growing body of work has begun to explore the intersection between social choice theory and RLHF~\citep{conitzer2024social, dai2024mapping, mishra2023ai, zhong2024provable, park2024principled, chakraborty2024maxmin, ge2024learning, swamy2024minimaximalist, siththaranjan2023distributional}, highlighting increased interest in this research direction. Several of these works advocate for incorporating axiomatic principles into the study of RLHF. In particular, three recent position papers~\citep{conitzer2024social, dai2024mapping, mishra2023ai} argue for the relevance of axiomatic frameworks in guiding RLHF design, aligning conceptually with our motivation, though they do not offer formal technical results.

Given that RLHF fails to satisfy key axioms from social choice theory, several recent works have proposed alternative approaches to address these shortcomings. \citet{ge2024axioms} introduced the lexical Copeland method to ensure pairwise majority consistency. \citet{maura2025jackpot} demonstrated that alignment via maximal lottery rule guarantees Condorcet consistency. Additionally, \citet{liu2025statistical} showed that Nash learning from human feedback can also satisfy Condorcet consistency. Our work differs from these approaches by showing that RLHF can, in fact, satisfy these properties under a practical and empirically motivated assumption.
\section{Conclusion}
\label{sec:conclusion}
This paper addresses that tension that prior work has shown that it fails to satisfy foundational axioms in social choice theory. We show that under a practical and mild assumption, the reward modeling component of RLHF satisfies both pairwise majority consistency and Condorcet consistency, explaining its effectiveness in real-world settings. Building on this insight, we show that a modified reward modeling objective that guarantees these desirable properties even under general preference profiles. Finally, to better align with the goals of generative models, we also introduce a new set of alignment desiderata tailored to LLMs. The main limitation of our work is that the modification objective requires significantly more labelers and data. Under current public datasets or labeling procedures, it is equivalent to standard RLHF. As a result, we leave further exploration to future work and hope that our findings will inform the development of more robust alignment methods in future LLM training pipelines.

\paragraph{Broader Impacts.} In Section~\ref{sec:axioms}, we introduce axioms aimed at faithfully preserving human preferences, which contribute positively to the fairness and representational equity of LLMs. As for potential negative impacts, this is primarily a theoretical study and does not pose direct societal risks.

\section*{Acknowledgments} 
This work was supported in part by NIH grant U01CA274576, ARPA-H Award D24AC00253, NSF grant DMS-2310679, a Meta Faculty Research Award, and Wharton AI for Business.

\newpage
\bibliographystyle{plainnat}
\bibliography{main.bib}
\newpage
\input{appendix}
\end{document}

%% file: appendix.tex
\appendix
\section{Proofs of Technical Results in Section \ref{sec:consistency}}
In this section, we provide the proofs of Theorem~\ref{thm:majorityandcondorcet}, Theorem~\ref{thm:majorityandcondorcet2}, and Lemma~\ref{lemma:copeland}. Since the proof of Theorem~\ref{thm:majorityandcondorcet} is a special case of Theorem~\ref{thm:majorityandcondorcet2}, and the latter builds upon Lemma~\ref{lemma:copeland}, we present the proofs in reverse order relative to their appearance in the main text.

\subsection{Proof of Lemma \ref{lemma:copeland}}
We consider the following general loss function used in reward modeling, which encompasses both the losses in Equation~\eqref{eq:rewardloss} and Equation~\eqref{eq:rewardloss2} for RLHF under pairwise comparisons:
\begin{equation}
\label{eq:rewardloss4}
\mathcal{L}(r_1,\ldots,r_n) = -\sum_{1\leq i<j\leq n} \left[ m_{ij} \log \sigma(r_i - r_j) + m_{ji} \log \sigma(r_j - r_i) \right],
\end{equation}
where \( \sigma(x) = \frac{1}{1 + e^{-x}} \) denotes the sigmoid function, and \( m_{ij} \) represents a general weight assigned to the loss when comparing $y_i$ to $y_j$. We now consider a general form of Lemma~\ref{lemma:copeland}:

\begin{lemma}
\label{lemma:generalrule}
Let \( m_{ij} \in \mathbb{R} \) with \( m_{ij} + m_{ji}=\mathtt{m} \), for any \( i, j \in [n] \). Define the score
\[
m_k = \sum_{j \ne k} \frac{m_{kj}}{\mathtt{m}}, \quad \text{for all } k \in [n].
\]
Then, the MLE of the objective in Equation~\eqref{eq:rewardloss4} satisfies, for all \( i, j \in [n] \),
\[
r_i > r_j \;\Leftrightarrow\; m_i > m_j.
\]
\end{lemma}
\begin{proof}
We want to compute the first-order condition with respect to \( r_k \). Since \( r_k \) appears in all terms where \( i = k \) or \( j = k \), we compute:
\[
\frac{\partial \mathcal{L}}{\partial r_k} 
= -\sum_{j \ne k} \left[ m_{kj} \cdot \frac{\partial}{\partial r_k} \log \sigma(r_k - r_j) 
+ m_{jk} \cdot \frac{\partial}{\partial r_k} \log \sigma(r_j - r_k) \right].
\]
Using the identities:
\[
\frac{d}{dx} \log \sigma(x) = 1 - \sigma(x), \quad 
\frac{d}{dx} \log \sigma(-x) = -\sigma(x),
\]
we get:
\[
\frac{\partial \mathcal{L}}{\partial r_k} 
= -\sum_{j \ne k} \left[ m_{kj} (1 - \sigma(r_k - r_j)) - m_{jk} \sigma(r_k - r_j) \right].
\]
This simplifies to:
\[
\frac{\partial \mathcal{L}}{\partial r_k} 
= -\sum_{j \ne k} \left[ m_{kj} - (m_{kj} + m_{jk}) \cdot \sigma(r_k - r_j) \right].
\]
Since \( \mathtt{m} = m_{kj} + m_{jk} \), the gradient with respect to \( r_k \) is given by:
\[
\frac{\partial \mathcal{L}}{\partial r_k} 
= -\sum_{j \ne k} \left[ m_{kj} - \mathtt{m} \cdot \sigma(r_k - r_j) \right].
\]
The first-order optimality condition sets this gradient to zero:
\[
\sum_{j \ne k} \left[ \frac{m_{kj}}{\mathtt{m}} - \sigma(r_k - r_j) \right] = 0,\ \forall k\in[n].
\]
By denoting
\[
m_k=\sum_{j \ne k}  \frac{m_{kj}}{\mathtt{m}},
\]
we have
\[
m_k=\sum_{j \ne k} \left[\sigma(r_k - r_j) \right],\ \forall k\in[n].
\]
Now, we consider the two equations of $r_i$ and $r_j$,
\[
m_i=\sum_{k \ne i} \left[ \sigma(r_i - r_k) \right]
\]
and
\[
m_j=\sum_{k \ne j} \left[ \sigma(r_j - r_k) \right].
\]
We subtract the second equation from the first and obtain
{\small\begin{equation*}
\begin{aligned}
&m_i-m_j\\ =&\sum_{k \ne i, k\ne j} \left[ \sigma(r_i - r_k)-\sigma(r_j - r_k) \right]+\sigma(r_i - r_j)-\sigma(r_j - r_i)\\
=&\sum_{k \ne i, k\ne j} \left[ \frac{\exp(r_i)}{\exp(r_i)+\exp(r_k)}-\frac{\exp(r_j)}{\exp(r_j)+\exp(r_k)} \right]+\frac{\exp(r_i)-\exp(r_j)}{\exp(r_i)+\exp(r_k)}\\
=&\sum_{k \ne i, k\ne j} \left[ \frac{\left(\exp(r_i)-\exp(r_j)\right)\exp(r_k)}{\left(\exp(r_i)+\exp(r_k)\right)\left(\exp(r_j)+\exp(r_k)\right)} \right]+\frac{\exp(r_i)-\exp(r_j)}{\exp(r_i)+\exp(r_k)}\\
=&\left(\exp(r_i)-\exp(r_j)\right)\left(\sum_{k \ne i, k\ne j} \left[ \frac{\exp(r_k)}{\left(\exp(r_i)+\exp(r_k)\right)\left(\exp(r_j)+\exp(r_k)\right)} \right]+\frac{1}{\exp(r_i)+\exp(r_k)}\right).
\end{aligned}
\end{equation*}}
Therefore, for all $i\ne j$,
\[
r_i > r_j \;\Leftrightarrow\; \exp(r_i)>\exp(r_j)\;\Leftrightarrow\;m_i > m_j.
\]
\end{proof}
Based on Lemma~\ref{lemma:generalrule}, the design of \( m_{ij} \) is essential for determining the ranking of \( r_1, r_2, \ldots, r_n \).
\paragraph{Copeland Rule.} In Equation~\eqref{eq:rewardloss2}, we have
\[
m_{ij} = \mathds{1}[\mathcal{P}(y_i \succ y_j) > 1/2].
\]
Then, \(\texttt{m} = 1\) (assuming no ties, e.g., an odd number of labelers or tie-breaking applied arbitrarily). Therefore,
\[
m_i = \sum_{k \neq i} \mathds{1}[\mathcal{P}(y_i \succ y_k) > 1/2],
\]
which corresponds exactly to the Copeland score \(w_i\). This completes the proof of Lemma~\ref{lemma:copeland}.
\paragraph{Borda Count Rule.} 
In addition to Lemma~\ref{lemma:copeland}, Lemma~\ref{lemma:generalrule} also implies that standard RLHF implements the Borda count rule. In standard RLHF, the pairwise count is defined as
\[
m_{ij} = \#(y_i \succ y_j),
\]
where \( \#(y_i \succ y_j) \) denotes the number of voters who prefer \( y_i \) over \( y_j \). In this case, \( \texttt{m} = m \), the total number of voters. Therefore,
\[
m_i = \sum_{k \neq i} \mathcal{P}(y_i \succ y_k),
\]
which corresponds exactly to the Borda count score \( \text{BC}_i \). Therefore, Lemma~\ref{lemma:generalrule}, as a more general result, reproduces the finding that standard RLHF implements the Borda count rule~\citep{siththaranjan2024distributional,maura2025jackpot}.

\subsection{Proof of Theorem \ref{thm:majorityandcondorcet2}}
By Lemma~\ref{lemma:copeland}, it suffices to show that the Copeland rule satisfies Pareto optimality, majority consistency, pairwise majority consistency, and Condorcet consistency.  
These properties are established in Theorems~\ref{thm:po}--\ref{thm:cc}, respectively.

\begin{theorem}[Pareto Optimality of the Copeland Rule]
\label{thm:po}
Let \(\mathcal{Y}=\{y_1,\dots,y_n\}\) be a set of candidates, and let each voter submit a strict ranking over \(\mathcal{Y}\).
The \textit{Copeland score} of candidate \(y_k\) is
\[
w_k \;=\; \#\bigl\{\,y_j\in \mathcal{Y}\setminus\{y_k\}\;:\; y_k \text{ is preferred to } y_j \text{ by a strict majority}\bigr\}.
\]
The Copeland rule ranks the candidates with the Copeland score.
The rule is Pareto optimal:  
if every voter strictly prefers candidate \(y\) to candidate \(y'\), then \(y'\) is never ranked above \(y\).
\end{theorem}

\begin{proof}
Assume every voter ranks \(y\) above \(y'\).
Fix any third candidate \(z\in \mathcal{Y}\setminus\{y,y'\}\).

For any individual ballot,
\[
\bigl(y' \succ z\bigr)\;\Longrightarrow\; \bigl(y \succ z\bigr),
\]
because \(y\) is above \(y'\) in that ballot.
Hence the set of voters who prefer \(y'\) to \(z\) is a \emph{subset} of those who prefer \(y\) to \(z\).
Consequently:
\[
\#\{\,\text{votes with } y' \succ z\,\}\;\le\;
\#\{\,\text{votes with } y \succ z\,\}.
\]

If a strict majority prefers \(y'\) to \(z\) (so \(y'\) defeats \(z\)), then the same strict majority (or more) prefers \(y\) to \(z\); therefore \(y\) also defeats \(z\).
Conversely, if \(y\) loses to \(z\), then \(y'\) must also lose to \(z\).
Thus, for every opponent \(z\),
\[
y' \text{ beats } z \;\;\Longrightarrow\;\; y \text{ beats } z,
\]
which implies
\[
w_{y'} \;\le\; w_y .
\]

Then, candidate \(y'\) can never obtain a Copeland score strictly greater than that of \(y\).  Therefore:

\begin{enumerate}
    \item \(y'\) is never ranked above \(y\) in the Copeland order.
    \item If the Copeland rule chooses a (possibly non-singleton) set of winners, \(y'\) can only be in that set if \(y\) is also in it (the case \(w_{y'} = w_y\)).
\end{enumerate}

Hence the Copeland rule satisfies the Pareto optimality criterion.
\end{proof}

\begin{theorem}[Majority Consistency of the Copeland Rule]
Let $\mathcal{Y}=\{y_1,\dots,y_n\}$ be a set of $n$ responses.  
Suppose a strict majority of voters rank some candidate $y^\star\in Y$ in first place.  
Then the Copeland rule---which selects the candidate(s) with the greatest number of pairwise majority wins---necessarily elects $y^\star$ (and no other candidate has a higher Copeland score).  Hence the Copeland rule satisfies the majority consistency criterion.
\end{theorem}

\begin{proof}
Because strictly more than half of all voters rank $y^\star$ first, those same voters prefer $y^\star$ to \emph{every} other candidate in every pairwise comparison.  Consequently, for every opponent $y_j\neq y^\star$ the majority outcome of the duel $(y^\star,y_j)$ is a win for $y^\star$.  Hence
\[
w_{y^\star}=n-1,
\]
where $w_i$ denotes the Copeland score (the number of pairwise majority wins) of candidate $y_i$.

Fix any other candidate $y_j\neq y^\star$.  
The contest $(y_j,y^\star)$ is a loss for $y_j$, so
\[
w_{y_j}\le n-2\;<\;w_{y^\star}.
\]

Thus $y^\star$ alone attains the maximal Copeland score, and the Copeland rule elects $y^\star$.  Therefore the rule is majority consistent.
\end{proof}

\begin{theorem}[Pairwise Majority Consistency of the Copeland Rule]
Let $\mathcal{Y}=\{y_1,\dots ,y_n\}$ be the set of responses, and let $\succ_M$ denote the (strict) pairwise majority relation.  
Assume that $\succ_M$ is \emph{complete} and \emph{transitive}; i.e., for every distinct $y_i,y_j\in \mathcal{Y}$ either $y_i\succ_M y_j$ or $y_j\succ_M y_i$, and no Condorcet cycles occur.  
Then the Copeland rule outputs the \emph{same linear ranking} as $\succ_M$.  Hence the Copeland rule is pairwise majority consistent.
\end{theorem}

\begin{proof}
Because $\succ_M$ is a complete, transitive order, we may relabel the candidates so that
\[
y_{(1)}\succ_M y_{(2)}\succ_M\cdots\succ_M y_{(n)} .
\]

For each candidate $y_{(k)}$, $k=1,\dots ,n$, let
\[
w_{(k)}\;=\;\#\bigl\{\,y_j\in \mathcal{Y}\setminus\{y_{(k)}\}:\;y_{(k)}\succ_M y_j \bigr\}
\]
be its Copeland score, i.e., the number of pairwise majority victories.

Because the majority order is linear: $y_{(1)}$ defeats every other candidate, so \(w_{(1)} = n-1\). $y_{(2)}$ loses only to $y_{(1)}$ and beats the remaining $n-2$ opponents, so \(w_{(2)} = n-2\). $y_{(k)}$ beats exactly the $n-k$ candidates ranked below it, so \(w_{(k)} = n-k\). $y_{(n)}$ defeats no one, so \(w_{(n)} = 0\).

Thus
\[
w_{(1)} > w_{(2)} > \cdots > w_{(n)} .
\]
Hence sorting candidates in \emph{decreasing} Copeland score yields precisely the order
\[
y_{(1)} \succ_M y_{(2)} \succ_M \cdots \succ_M y_{(n)} .
\]

The Copeland rule ranks all candidates in the identical linear order given by the pairwise majority relation whenever that relation is transitive.  Therefore the Copeland rule satisfies pairwise majority consistency.
\end{proof}

\begin{theorem}[Condorcet Consistency of the Copeland Rule]
\label{thm:cc}
Let $\mathcal{Y}=\{y_1,\dots,y_n\}$ be a set of $n$ candidates and let each voter provide a strict (complete, transitive) ranking over $Y$.  
For every candidate \(y_i\in Y\), define its Copeland score
\[
w_i \;=\;\#\bigl\{\,y_j\in \mathcal{Y}\setminus\{y_i\}\;:\; y_i \text{ is preferred to } y_j \text{ by a strict majority}\bigr\}.
\]
The Copeland rule selects the candidate(s) with the maximal Copeland score.  
This rule is Condorcet consistent: whenever the profile admits a Condorcet winner, the Copeland rule elects (and only elects) that candidate.
\end{theorem}

\begin{proof}
Assume the profile has a Condorcet winner, say $y_c\in \mathcal{Y}$.  
By definition, $y_c$ defeats every other candidate in a pairwise majority comparison.  Hence
\[
w_c \;=\; n-1.
\]

Now fix any other candidate $y_j\neq y_c$.  
Because $y_c$ beats $y_j$, the pairwise contest $(y_j,y_c)$ counts as a \emph{loss} for $y_j$.  Consequently
\[
w_j \;\le\; n-2
\quad\Longrightarrow\quad
w_j \;<\; w_c .
\]

Therefore \(w_c\) is strictly greater than the Copeland score of every rival.  The Copeland rule, which chooses all candidates with the highest score, selects \(y_c\) (uniquely).  Thus, whenever a Condorcet winner exists, the Copeland rule elects that candidate and no other, establishing Condorcet consistency.
\end{proof}
\subsection{Proof of Theorem \ref{thm:majorityandcondorcet}}
Under Assumption~\ref{ass:one_labeler}, pairwise-majority consistency and Condorcet consistency remain well defined; hence their proofs are special cases of Theorem~\ref{thm:majorityandcondorcet2}.  

By contrast, the same assumption allows the preference profile to be incomplete, so Pareto optimality and majority consistency are not well defined.

\section{Proofs of Technical Results in Section \ref{sec:axioms}}
\subsection{Proof of Theorem \ref{thm:pm}}
\begin{proof}
\textbf{Necessary.}
Assume a preference matching distribution
$\boldsymbol{p}^{\star}$ exists.  
Define $r_i:=\log p_i^{\star}$ for every $i$.

Then,
\[
  \mathcal{P}(y_i \succ y_j)
  \;=\;
  \frac{p_i^{\star}}{p_i^{\star}+p_j^{\star}}
  \;=\;
  \frac{\exp(r_i)}{\exp(r_i)+\exp(r_j)}
  \quad\text{for all }i\neq j,
\]
which is exactly the BT form with parameters
$(r_1,\dots ,r_n)$.  Hence the profile is BT‐embeddable.

\medskip\noindent
\textbf{Sufficiency.}
Conversely, suppose the profile is BT‐embeddable;  
let $r_1,\dots ,r_n$ be BT parameters satisfying
\[
  \mathcal{P}(y_i \succ y_j)
  \;=\;
  \frac{\exp(r_i)}{\exp(r_i)+\exp(r_j)}
  \quad\text{for all }i\neq j .
\]
Set $p_i^{\star}:=\exp(r_i)$ and normalise, if desired, so that
$\sum_{k}p_k^{\star}=1$.  Then
\[
  \frac{p_i^{\star}}{p_i^{\star}+p_j^{\star}}
  \;=\;
  \frac{\exp(r_i)}{\exp(r_i)+\exp(r_j)}
  \;=\;
  \mathcal{P}(y_i \succ y_j),
\]
so $\boldsymbol{p}^{\star}$ satisfies the requirement. Therefore, a preference matching
distribution exists.
\end{proof}
\subsection{Proof of Theorem~\ref{thm:rlhf_pe}}
\begin{proof}
Let $\mathcal{L}(r_1,\dots ,r_n)$ be the reward learning objective:
\[
\mathcal{L}(r_1,\ldots ,r_n)
=
-\sum_{1\le i<j\le n}
\left[
m_{ij}\log\sigma(r_i - r_j)
+
m_{ji}\log\sigma(r_j - r_i)
\right],
\qquad
\sigma(x) = \frac{1}{1 + e^{-x}},
\]
where $m_{ij}$ is the number of labelers who prefer $y_i$ over $y_j$.
Let $\hat{\boldsymbol{r}} = (\hat{r}_1,\dots,\hat{r}_n)$ denote the MLE of this objective, and define $p_i^\star \propto \exp(\hat{r}_i)$.

Suppose $y_i$ and $y_j$ are equally preferred, and let $y_{(1)}, y_{(2)}, \dots, y_{(n)}$ be a relabeling of $y_1, \dots, y_n$ that permutes candidates so that $y_j$ is mapped to the position of $y_i$, i.e., $y_j \mapsto y_{(i)}$. Let $\boldsymbol{\tau}'$ be the transformed profile, which is equivalent to $\boldsymbol{\tau}$ under relabeling.

By the definition of Preference Equivalence, the transformed pairwise counts satisfy:
\[
m_{(i)(k)} = m_{ik} \quad \text{for all } k \ne i,
\]
and since $y_j$ maps to position $(i)$ under the relabeling, we also have:
\[
m_{j(k)} = m_{ik} \quad \text{for all } k \ne i.
\]
Therefore,
\[
m_j = \sum_{k \ne j} m_{j(k)} = \sum_{k \ne i} m_{ik} = m_i.
\]

Since the aggregated weights determining $r_j$ and $r_i$ are equal, it follows that $\hat{r}_i = \hat{r}_j$ under the MLE solution.

\paragraph{Equal probabilities.}
The aggregation rule assigns
\[
p_k^\star \propto \exp(\hat{r}_k),
\]
so $\hat{r}_i = \hat{r}_j$ implies $p_i^\star = p_j^\star$ after normalization. Thus, the reward-modeling aggregation rule satisfies the Preference Equivalence axiom.
\end{proof}

\subsection{Proof of Theorem~\ref{thm:exist}}
\begin{proof}
The main idea is that any individual strict ranking is BT-embeddable. Suppose a voter expresses the preference
\[
y_{(1)} \succ y_{(2)} \succ \cdots \succ y_{(n)}.
\]
This is equivalent to assuming that the pairwise comparison probabilities satisfy
\[
\mathcal{P}\bigl(y_{(i)} \succ y_{(j)}\bigr) = 1
\qquad \text{for all } i < j,\ \text{with } i,j \in \{1, \dots, n\}.
\]
To reflect the relative distances between responses in a BT-compatible form, we may approximate these deterministic comparisons by
\[
\mathcal{P}\bigl(y_{(i)} \succ y_{(i+1)}\bigr) = 1 - \epsilon
\qquad \text{for all } i \in \{1, \dots, n-1\},
\]
with \(\epsilon > 0\) small. Then, by the Bradley--Terry formulation, we set the ratios
\[
\frac{p_i^*}{p_{i+1}^*} = \frac{1 - \epsilon}{\epsilon},
\]
which leads to a closed-form preference-matching distribution consistent with the BT model.
let the adjacent pairwise probabilities satisfy  
\[
\mathcal{P}\!\bigl(y_{(i)}\succ y_{(i+1)}\bigr)=1-\epsilon ,
\qquad i=1,\dots ,n-1,
\quad\text{with } \epsilon\in(0,\tfrac12)\ \text{and }\ \epsilon\to0.
\]

Under a Bradley--Terry–type representation we require  
\[
\frac{p_i^{\ast}}{p_i^{\ast}+p_{i+1}^{\ast}} = 1-\epsilon
\quad\Longleftrightarrow\quad
\frac{p_{i+1}^{\ast}}{p_i^{\ast}} = \frac{\epsilon}{1-\epsilon}
\quad(i=1,\dots ,n-1).
\]
Define the constant
\[
c \;=\; \frac{\epsilon}{1-\epsilon}\in(0,1),
\]
so that the adjacent ratios are
\(
p_{i+1}^{\ast}=c\,p_{i}^{\ast}
\).
Iterating,
\[
p_i^{\ast} = c^{\,i-1}\,p_1^{\ast},
\qquad i=1,\dots ,n.
\]

\paragraph{Normalisation.}
Impose \(\sum_{i=1}^{n}p_i^{\ast}=1\):
\[
p_1^{\ast}\sum_{k=0}^{n-1}c^{k}=1
\quad\Longrightarrow\quad
p_1^{\ast} = \frac{1-c}{1-c^{\,n}} .
\]
Hence the closed-form solution is
\[
p_i^{\ast} \;=\; \frac{(1-c)\,c^{\,i-1}}{1-c^{\,n}},
\qquad
c=\dfrac{\epsilon}{1-\epsilon}\in(0,1).
\;
\]
This shows that the preference matching distribution $\boldsymbol{p^*}$ exists, thus the individual preference is BT-embeddable. 

Then, by letting the \(i\)-th subgroup consist solely of the \(i\)-th voter, we obtain a partition of \(\mathcal{U}\) such that each subgroup is BT-embeddable.

\paragraph{Limit as \(\epsilon\to0\).}
In addition, since \(c=\epsilon/(1-\epsilon)\to0\),
\[
p_1^{\ast}\;\longrightarrow\;1,
\qquad
p_i^{\ast}\;\longrightarrow\;0 \quad(i\ge2),
\]
so the distribution concentrates on the top-ranked response \(y_{(1)}\), as expected when each adjacent comparison is almost surely won by the higher candidate.
\end{proof}
\subsection{Proof of Theorem \ref{thm:unique}}
\begin{proof}
Let \(\mathcal{U}_k\) be a subgroup of $\mathcal{U}$ with \(m_k\) voters, where each individual voter \(i \in \{1, \ldots, m_k\}\) has a preference matching distribution \(\boldsymbol{p}_i^*\).  
If the group \(\mathcal{U}_k\) is BT-embeddable, then the preference-matching distribution is given by
\[
\boldsymbol{p}^*_{\mathcal{U}_k} = \frac{1}{m_k} \sum_{i=1}^{m_k} \boldsymbol{p}_i^*.
\]
Then, for any partition of $\mathcal{U}$,
\begin{equation*}
    \boldsymbol{p}_{\mathcal{U}}^* = \sum_{k=1}^K \frac{|\mathcal{U}_k|}{|\mathcal{U}|} \, \boldsymbol{p}^*_k= \sum_{k=1}^K \frac{m_k}{m} \, \boldsymbol{p}^*_k=\frac{1}{m} \sum_{i=1}^{m} \boldsymbol{p}_i^*.
\end{equation*}
This expression is independent of the specific partition, and therefore the group preference matching distribution is unique.
\end{proof}
\paragraph{Estimiation of $\boldsymbol{p^*}$.} 
Since \(c=\epsilon/(1-\epsilon)\to0\),
\[
p_1^{\ast}\;\longrightarrow\;1,
\qquad
p_i^{\ast}\;\longrightarrow\;0 \quad(i\ge2).
\]
We have
$p_i^* \rightarrow \#(y_i\ \text{ranked first}) / m$
when $\epsilon\rightarrow 0$.
\section{Analysis of Policy optimization of RLHF}
We begin by briefly introducing Step 3 of RLHF.
\paragraph{Step 3: Policy Optimization and RLHF Fine-tuning.} Let $\pi_{\phi}(y|x)$ be the probability distribution of the responses given a prompt $x$, where $\phi$ denotes the weights of the LLM. The goal of (unregularized) RLHF fine-tuning is to maximize the expected reward, i.e., $\max_\phi \mathbb{E}_{y\sim\pi_{\phi}(\cdot|x)} r(x,y)$. To mitigate over-optimization of the reward model, additional regularization should be added to the objective function, i.e., $\max_\phi \mathbb{E}_{y\sim\pi_{\phi}(\cdot|x)} r(x,y)-\text{Regularizer},$ where, unless specified otherwise, the expectation is taken over the randomness of $x$ and, conditional on $x$, the randomness of $y$ sampled from $\pi_{\phi}(\cdot|x)$. In practice, $x$ can be either sampled from a fixed database of prompts \citep{rafailov2023direct} or adaptively based on prior responses \citep{xiong2023iterative}. RLHF uses a KL penalty between the RLHF model and the reference model, which is the pretrained or SFT model, at each token as the regularizer. The loss function of regularized RLHF fine-tuning is
\begin{equation}\label{eq:prlloss}
\max_\phi \mathbb{E}_{y\sim\pi_{\phi}(\cdot|x)} r(x,y)-\beta D_{\text{KL}}(\pi_\phi (y|x)\| \pi_{\textnormal{ref}}(y|x)),
\end{equation}
where $\beta > 0$ is a parameter controlling the deviation from the base reference policy $\pi_{\textnormal{ref}}$, namely the initial SFT model $\pi_{\text{SFT}}$. Above, the Kullback--Leibler divergence between $\pi_\phi (y|x)$ and $\pi_{\textnormal{ref}}(y|x)$ is defined to be $\mathbb{E}_{y\sim\pi_{\phi}(\cdot|x)}\big[\log (\pi_\phi (y|x))-\log (\pi_{\textnormal{ref}} (y|x))\big]$.
In practice, $\pi_\phi (y|x)$ is calculated token-wise, let $y=[T_1,\cdots,T_L]$, it can be expressed in
\begin{equation*}
\log (\pi_\phi (y|x))=\sum_{i=1}^L\bigg[ \log\pi_\phi(T_i\mid T_{i-1},\cdots,T_1,T_0)\bigg],
\end{equation*}
where we further denote $T_0=x$. The same notation is used for $\pi_{\textnormal{ref}}(y|x)$. For clarity, the term RLHF problem specifically refers to the problem defined in \eqref{eq:prlloss}. When comparing various RLHF approaches, \eqref{eq:prlloss} is referred to as either standard RLHF or KL RLHF.

\paragraph{Policy Optimization as an Aggregation Rule.}
Based on the RLHF fine-tuning objective, which takes the comparison dataset \( D_{\textnormal{comp}} \) as input, the resulting fine-tuned LLM defines an induced distribution \(\pi(y_i \mid x)\) over the responses \(y_1, \ldots, y_n\). This distribution induces a strict total ranking over the responses.

\paragraph{Policy Optimization Preserves the Winner.}
The objective of RLHF is to maximize the output probability of the response with the highest reward. Therefore, when the regularization parameter \(\beta\) is sufficiently small—such that no other response can surpass the (regularized) reward of the original top-ranked response—policy optimization will preserve the winner selected by the reward function. That is, the response with the highest reward remains the most probable under the optimized policy.

\paragraph{Policy Optimization as a Probabilistic Aggregation Rule.}
The fine-tuned LLM defines an induced distribution \(\pi(y_i \mid x)\) over the responses \(y_1, \ldots, y_n\), which naturally serves as a \emph{probabilistic aggregation rule}---that is, a probability distribution over responses consistent with the learned preference structure.

\paragraph{PM RLHF.}
Unfortunately, standard RLHF does not necessarily preserve the (group) preference-matching distribution \(\boldsymbol{p}^*\). As discussed in \citet{xiao2024algorithmic}, applying an entropy-regularized objective of the form
\[
\max_\pi \; \mathbb{E}_{y \sim \pi} [r(x,y)] + H(\pi),
\]
with entropy regularization coefficient set to one, yields the optimal solution
\[
\pi(y_i \mid x) \propto \exp(r(x, y_i)),
\]
which recovers the BT distribution parameterized by the reward function \(r\).

We refer readers to \citet{xiao2024algorithmic} for a detailed discussion. This insight suggests that, with properly designed reward modeling and entropy-regularized policy optimization, it is possible to align the policy distribution \(\pi\) with the group preference-matching distribution \(\boldsymbol{p}^*\).

\section{Other Related Work}
Several methods have been developed to improve preference fine-tuning in RLHF. \citet{li2023remax} demonstrated that proximal policy optimization (PPO)~\citep{schulman2017proximal} may not fully leverage RLHF’s capacity to align large language models (LLMs) with human preferences. The work by \citet{tang2024understanding} explored discrepancies between online and offline learning paradigms for alignment, while \citet{ye2024theoretical} proposed an online iterative RLHF method based on a general preference structure. Additionally, \citet{li2025preserving} investigated how model diversity can be maintained during the supervised fine-tuning (SFT) stage. Outside of fine-tuning, model editing has emerged as a complementary strategy to modify LLM behavior across tasks \citep{jin2025finetuning}.

Several extensions of the DPO method have been introduced to address its limitations \citep{liu2023statistical,azar2024general,chang2024dataset,gorbatovski2024learn,rafailov2024r,yang2024asymptotics}. Nevertheless, recent analyses indicate that DPO may underperform compared to reward-optimized RLHF methods in terms of alignment effectiveness \citep{li2023policy,xu2024dpo,tajwar2024preference}. This performance gap has been partly attributed to representational mismatches inherent in DPO, as discussed by \citet{li2023policy} and \citet{xu2024dpo}, which constrain its robustness relative to reinforcement learning techniques like PPO. Furthermore, the on-policy nature of reward-based fine-tuning helps alleviate distribution shift between training and deployment scenarios, thereby boosting LLM reliability \citep{tajwar2024preference}. The calibration challenges introduced by RLHF have also been studied by \citet{xiao2025restoring}.

An alternative line of work introduces Nash learning from human feedback (NLHF), which aims to align LLMs under general preference models \citep{munos2023nash,liu2025statistical,shi2025fundamental}. In this direction, \citet{wang2025magnetic} examined the convergence behavior of NLHF and showed that the last iterate converges to a Nash equilibrium.

\section{Group Preference Matching Distribution and Mixture of PL Model}
\paragraph{Mixture of Plackett--Luce Models.}
The Plackett--Luce (PL) model is a widely used probabilistic model for ranking data.  
Given a set of \(n\) candidates \(Y = \{y_1, \dots, y_n\}\), the PL model assigns a probability to each possible ranking \(\sigma\) based on a parameter vector \(\boldsymbol{\theta} = (\theta_1, \dots, \theta_n)\), where \(\theta_i > 0\) represents the latent utility of candidate \(y_i\). The probability of ranking \(\sigma = (\sigma(1), \dots, \sigma(n))\) is
\[
\mathcal{P}_{\text{PL}}(\sigma \mid \boldsymbol{\theta}) = \prod_{i=1}^{n} \frac{\theta_{\sigma(i)}}{\sum_{j=i}^{n} \theta_{\sigma(j)}}.
\]

To model heterogeneity across subpopulations or latent preference types, one may use a \emph{mixture of PL models}. In this setting, the observed rankings are assumed to be generated from one of \(K\) latent components, each with its own PL parameter vector \(\boldsymbol{\theta}^{(k)}\) and associated mixture weight \(\pi_k\), where \(\pi_k \geq 0\) and \(\sum_{k=1}^{K} \pi_k = 1\). The mixture model defines the ranking probability as
\[
\mathcal{P}_{\text{Mix-PL}}(\sigma) = \sum_{k=1}^{K} \pi_k\, \mathcal{P}_{\text{PL}}(\sigma \mid \boldsymbol{\theta}^{(k)}).
\]

This model captures diverse preference patterns across groups and is especially useful when the population consists of multiple subgroups with distinct but internally consistent rankings.

\paragraph{Comparison with Mixture of BT/PL Models.}
The group preference matching distribution (GPMD), as defined in Definition~\ref{def:gpmd}, aggregates subgroup-level preference-matching distributions:
\[
\boldsymbol{p}^* = \sum_{k=1}^K \frac{|\mathcal{U}_k|}{|\mathcal{U}|} \, \boldsymbol{p}^*_k,
\]
where each subgroup \(\mathcal{U}_k\) has a BT-embeddable preference profile and a corresponding preference-matching distribution \(\boldsymbol{p}^*_k\). This formulation resembles the mixture form commonly used in \emph{mixture of BT or PL models}, where the overall ranking distribution is expressed as a convex combination of component-wise BT (or PL) models.

Despite the formal similarity in expression, the underlying semantics are fundamentally different:

\begin{itemize}
    \item \textbf{Group Preference Matching Distribution (GPMD):} 
    The distribution \(\boldsymbol{p}^*\) is a deterministic aggregation of observed human preference profiles. Each \(\boldsymbol{p}^*_k\) directly reflects the empirical ranking behavior within subgroup \(\mathcal{U}_k\), without assuming a generative model. The GPMD thus serves as an interpretable, population-level summary of heterogeneous but internally consistent (BT-embeddable) human preferences.

    \item \textbf{Mixture of BT/PL Models:}
    In contrast, a mixture model assumes that each observed ranking is generated from a latent BT or PL component chosen according to a categorical distribution. The mixture weights and parameters are typically estimated via maximum likelihood (e.g., EM algorithm). This approach reflects a probabilistic generative view and introduces latent structure not directly tied to explicit subgroupings in the data.

\end{itemize}

In short, the GPMD is an \emph{explicit aggregation of empirical preference structures}, while mixture models reflect an \emph{implicit generative assumption} with unobserved component memberships. The former is rooted in descriptive modeling of human feedback; the latter in statistical inference over latent preference types.